\documentclass[journal]{IEEEtran}
\usepackage{amsmath, amsfonts, amssymb}
\usepackage{mathrsfs}
\usepackage[ruled,linesnumbered]{algorithm2e} 

\usepackage{graphicx}
\usepackage{multirow}
\graphicspath{{images/}}

\usepackage{booktabs}

\usepackage[square,sort&compress,numbers]{natbib}

\usepackage{url}

\usepackage[utf8]{inputenc}

\begin{document}

\newcommand{\etal}{\textit{et al}.}

%

% ******************************paper title****************************** 
% \title{ImMPI: Novel View Synthesis in Remote Sensing with Implicit MPI Representation}
\title{Remote Sensing Novel View Synthesis with Implicit Multiplane Representations}

% ******************************author names****************************** 
\author{Yongchang Wu, Zhengxia Zou$^\star$ and Zhenwei Shi, 
\IEEEmembership{Member,~IEEE}

\thanks{The work was supported by the National Natural Science Foundation of China under the Grant 62125102. \emph{(Corresponding author: Zhengxia Zou (e-mail: zhengxiazou@buaa.edu.cn))}}

\thanks{Yongchang Wu and Zhenwei Shi are with the Image Processing Center, School of Astronautics, Beihang University, Beijing 100191, China, and with the Beijing Key Laboratory of Digital Media, Beihang University, Beijing 100191, China, and also with the State Key Laboratory of Virtual Reality Technology and Systems, School of Astronautics, Beihang University, Beijing 100191, China.
Zhengxia Zou is with Department of Guidance, Navigation and Control, School of Astronautics, Beihang University, Beijing 100191, China.}% <-this % stops a space
% <-this % stops a space
}

% ******************************The paper headers****************************** 


% make the title aread
\maketitle

% ******************************abstract******************************
\begin{abstract}

Novel view synthesis of remote sensing scenes is of great significance for scene visualization, human-computer interaction, and various downstream applications. Despite the recent advances in computer graphics and photogrammetry technology, generating novel views is still challenging particularly for remote sensing images due to its high complexity, view sparsity and limited view-perspective variations. In this paper, we propose a novel remote sensing view synthesis method by leveraging the recent advances in implicit neural representations. Considering the overhead and far depth imaging of remote sensing images, we represent the 3D space by combining implicit multiplane images (MPI) representation and deep neural networks. The 3D scene is reconstructed under a self-supervised optimization paradigm through a differentiable multiplane renderer with multi-view input constraints. Images from any novel views thus can be freely rendered on the basis of the reconstructed model. As a by-product, the depth maps corresponding to the given viewpoint can be generated along with the rendering output. We refer to our method as Implicit Multiplane Images (\textbf{ImMPI}). To further improve the view synthesis under sparse-view inputs, we explore the learning-based initialization of remote sensing 3D scenes and proposed a neural network based Prior extractor to accelerate the optimization process. In addition, we propose a new dataset for remote sensing novel view synthesis with multi-view real-world google earth images. Extensive experiments demonstrate the superiority of the ImMPI over previous state-of-the-art methods in terms of reconstruction accuracy, visual fidelity, and time efficiency. Ablation experiments also suggest the effectiveness of our methodology design. Our dataset and code can be found at \url{https://github.com/wyc-Chang/ImMPI}

% and validate the effectiveness and efficiency of our algorithm. 

% Benefiting from the advancement of sensors and the development of photogrammetry technology, 3D reconstruction and visualization of remote sensing scenes have achieved good results during recent years. However, generating novel views from multi-view input of specific scenes is still a challenging problem. 
% The sparsity of earth observation perspective and large scale of scenes make it difficult in remote sensing. 

% we propose a learning-based method to extract the distribution prior of scene contents, which serve as an ImMPI initialization method to accelerate the optimization process.

% We present \textbf{ImMPI} to present 3D space combining implicit neural network and explicit multiplane images (MPI) representation. 

% In order to improve the accuracy of scene reconstruction with sparse perspectives, we propose a learning-based method to extract the distribution prior of scene contents, which serve as an ImMPI initialization method to accelerate the optimization process. 

% has less artifacts in rendered image and performs better than existing methods. 

\end{abstract}

% ****************************** keywords****************************** 
% Note that keywords are not normally used for perreview papers.

\IEEEpeerreviewmaketitle

%  ******************************Introduction******************************

\section{Introduction}
\label{sec:intro}
\IEEEPARstart{N}{ovel} view synthesis aims at rendering novel images of a 3D scene from arbitrary query viewpoints given a set of pre-collected multi-view images as input. In remote sensing, novel view synthesis has substantial application potential for various tasks such as 3D scene reconstruction, urban management and disaster assessment.

% Despite its usefulness, the novel view synthesis problem is challenging in remote sensing due to the complexity of the scene, especially large scale and large depth range.

% These methods . 

% based on can accurately reconstruct small-scale scenes especially at object level. 

% However, traditional 3D representations have limitations to reconstruct remote sensing scenes. Mesh and volume require considerable memory to store the geometry information of 3D space, which limits the application in large-scale scenes. 

% Additionally, the images rendered by these methods may suffer from severe artifacts in occlusion areas. Therefore, it is of great significance to build a new 3D scene representation that can take into account the characteristics of remote sensing scenes.

\begin{figure}
    \centering
    \includegraphics[width=0.5\textwidth]{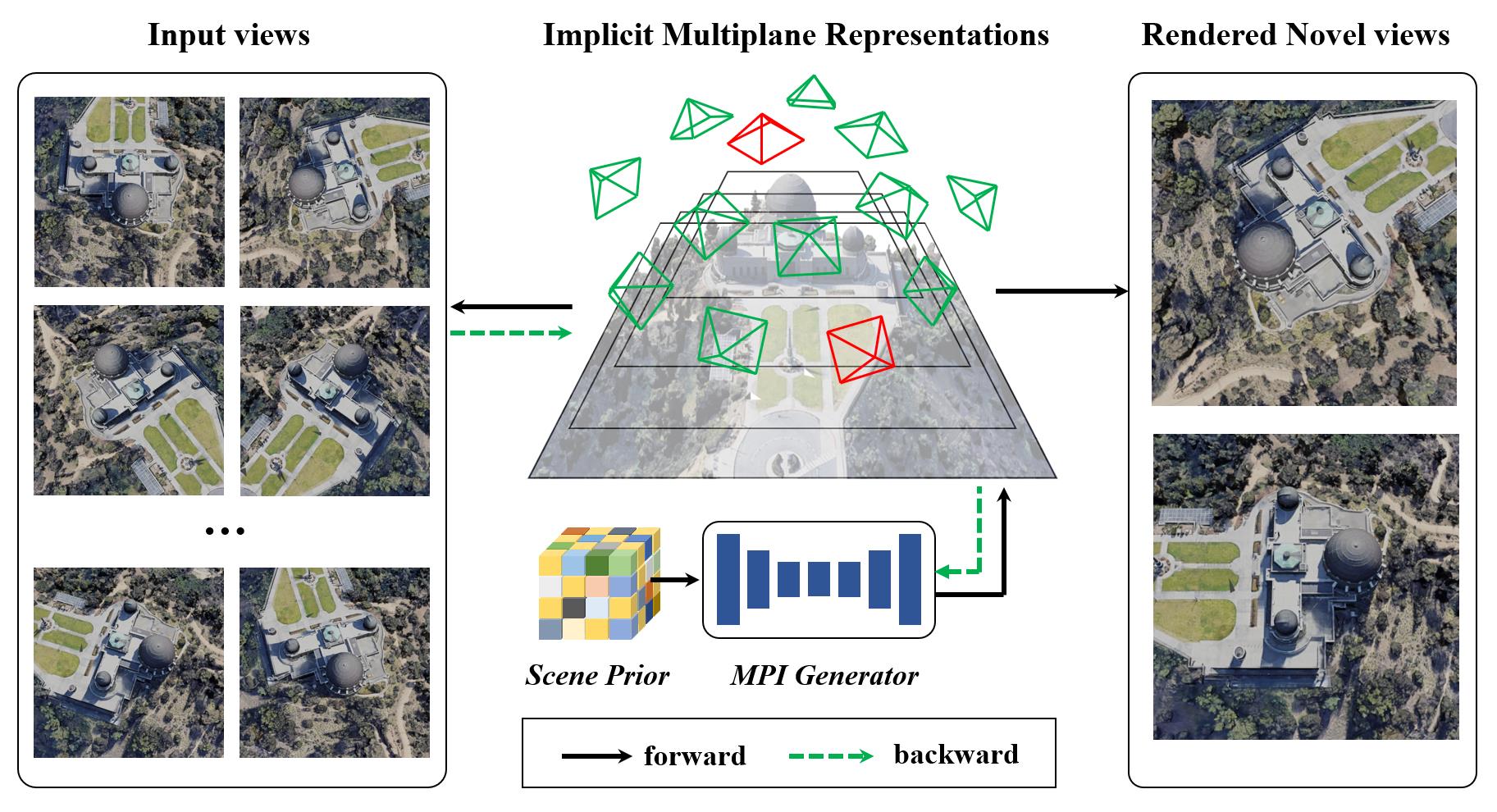}
    \caption{We propose a new method for remote sensing novel view synthesis. Unlike recent approaches designed based on volume rendering, our method takes advantage of implicit multiplane images  (ImMPI) representation to generate novel views from sparse posed images. In this figure, we visualize 11 training views (marked as green) and 2 novel views (marked as red) poses and show corresponding images rendered from optimized ImMPI.}
    \label{fig:teaser}
\end{figure}

Generating novel views from multi-view remote sensing images is challenging due to the high complexity, view-sparsity and limited view-perspective variations of remote sensing scenes. Recent approaches to novel view synthesis and 3D scene construction are usually designed based on mesh rendering~\cite{kanazawa2018learning,wang2018pixel2mesh,groueix2018papier} and volumetric representation~\cite{brock2016generative,wu20153d, flynn2019deepview, kar2017learning, tulsiani2017multi} techniques, mainly focusing on small-scale scenes, particularly at object level. Recently, implicit neural representation~\cite{chen2019learning, genova2019learning, genova2020local, jiang2020local, mescheder2019occupancy, park2019deepsdf}, as an emerging technique in computer vision and graphics has brought great attention to novel view synthesis and 3D representations. Implicit neural representation provides a novel way to parameterize continuous differentiable signals with neural networks, including volumes and radiance signals in 3D scenes. Based on implicit neural representation, many approaches have been proposed for novel view synthesis very recently. NeRF~\cite{mildenhall2020nerf} is known as a representative of such group of approaches. NeRF is proposed to encode the 3D shapes into the network weights, combined with differentiable rendering to achieve end-to-end optimization, where the inefficiency of scene representation and the complexity of rendering are significantly reduced. CityNeRF~\cite{xiangli2021citynerf} extends the NeRF from object-level to city-scale with multi-scale remote sensing images as input. A progressive training paradigm is proposed in CityNeRF to store scene details by gradually adding network modules. However, the implicit neural representation in these methods is still limited by the traditional volume rendering process, and thus may suffer from a slow rendering speed. 

In remote sensing image view synthesis, restricted by the camera movement of the mounted platform (e.g., satellites and UAVs), the camera usually has a limited range of perspective variation as it flies over the scene. In addition, when the collected views are very sparse, it will become more difficult to reconstruct the 3D scene accurately. Recent methods like SinSyn~\cite{2019SynSin} and MINE~\cite{li2021mine} explore to generate novel view from single image input. However, due to the absence of real scale, it is difficult for the above discussed methods to render high-quality images in remote sensing applications. 

To tackle the above challenges, we propose a new method called Implicit Multiplane Images (ImMPI) for remote sensing novel view synthesis. We incorporate multiplane images, an explicit representation naturally suitable for remote sensing with the recent advances in implicit neural representation. In the proposed method, the 3D scene is constructed under a self-supervised optimization paradigm through a differentiable multiplane renderer with multi-view input constraints. Images from any novel view can be thus freely rendered based on the reconstructed 3D scene model. As a by-product, the depth maps corresponding to the given viewpoint can be generated along with the rendering output. In addition, an initialization method is proposed with the motivation that 3D scene priors learned from large remote sensing datasets can be applied across scenes, which further improves the optimization stability and efficiency under sparse-view inputs. Since there are no publicly available datasets for remote sensing novel view synthesis, we build a new dataset for this task using real-world google earth images. Extensive experiments demonstrate the superiority of the ImMPI over previous state-of-the-art methods in terms of reconstruction accuracy, visual fidelity and time efficiency. 

The contribution of our work can be summarised as follows:

\begin{itemize}
\item We propose implicit MPI representation (ImMPI), a novel method to represent remote sensing 3D scenes. Combining the advantages of implicit neural representation and explicit MPI, the proposed method is naturally suitable for remote sensing novel view synthesis and enables fast rendering.

\item We introduce a learning-based network for ImMPI initialization. By extracting 3D scene distribution priors, the optimization process can be significantly accelerated and stabilized.

\item We introduce a new dataset for remote sensing novel view synthesis. The dataset consists of 16 real-world 3D scenes collected from Google Earth as well as their multi-view images, including mountains, urban area, buildings, parks, villages, etc. We also made our code publicly available. The dataset and code can be found at \url{https://github.com/wyc-Chang/ImMPI}.

\end{itemize}

% swings within a small range in earth observation with the optical axis almost perpendicular to the ground. 
% Therefore, the perspectives provided in remote sensing are very sparse, which makes it difficult to accurately reconstruct the 3D scene. 

% How to synthesize novel views based on sparse perspectives in remote sensing is another difficulty.

% that parameterizes a continuous differentiable signal with a neural network  alternative method brings inspiration to novel view synthesis.

\begin{figure*}[t]
    \centering
    \includegraphics[width=\textwidth]{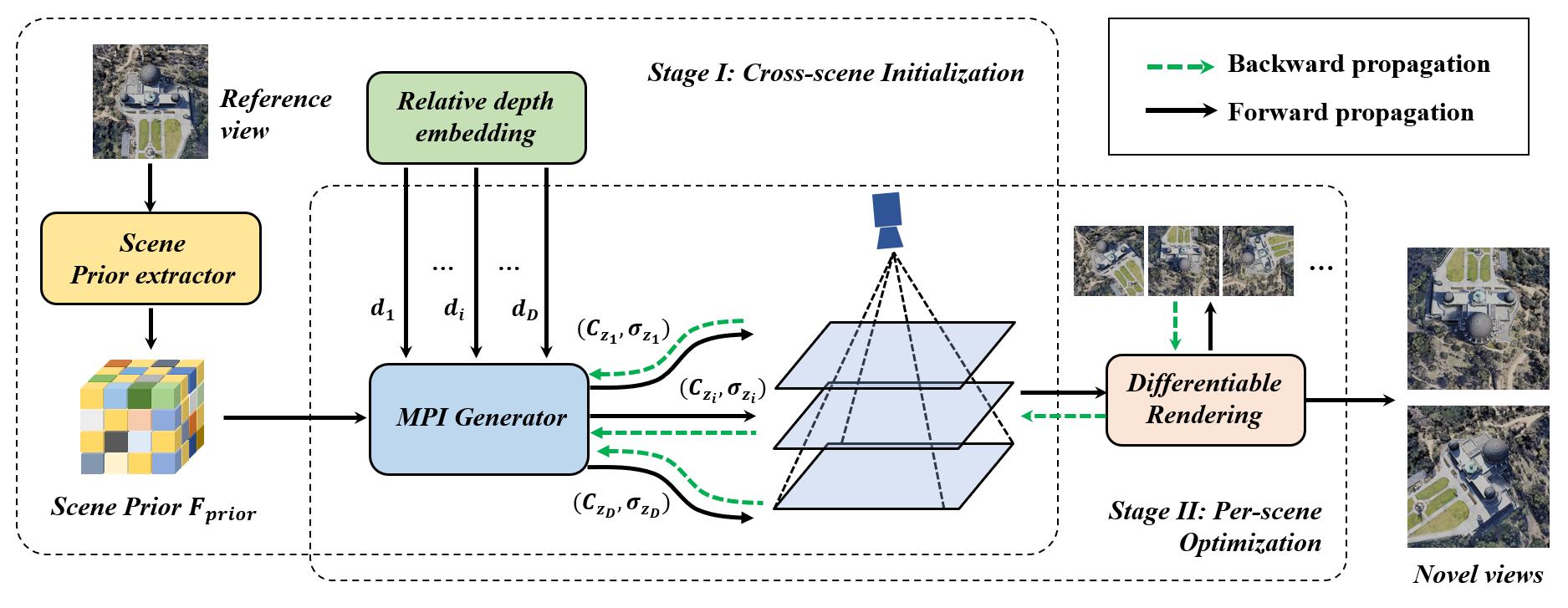}
    \caption{An overview of our method. The processing pipeline of the proposed method consists of two stages. 1) Cross-scene initialization: Given a selected perspective denoted as reference view, we propose a Prior extractor encoder that produces the latent feature $F_{prior}$ as the initialization of implicit MPI. The encoder is trained on different remote sensing scenes in a self-supervised way. 2) Per-scene optimization: Given a set of input views, we iteratively optimize the ImMPI representation and then render novel views from optimized ImMPI by differentiable rendering.}
    \label{fig:overall}
\end{figure*}

%  ******************************Related Work****************************** 
\section{Related Work}

\subsection{3D Representation for novel view synthesis}

Novel view synthesis aims at rendering unobserved viewpoints from a scene given a number of images and camera poses as inputs. It can be modeled as a two-stage process where the first stage recovers geometry from multi-view images and the second one renders images corresponding to given viewpoints. The representation quality of 3D scene is crucial to the quality of the rendered novel views. In this subsection, we introduce common scene representation methods especially for novel view synthesis task, including explicit representations and implicit representations.

Explicit 3D scene representation includes optical flow, mesh, volume, etc. Some early approaches~\cite{levoy1996light,gortler1996lumigraph,davis2012unstructured} reconstruct the optical flow field from multi-view images and achieve view synthesis by interpolation. However, these methods require very dense input views of the scene, which limits their application scope.
Some recent methods explore mesh-based representations for novel view synthesis. Liu \etal proposed mesh-based novel view synthesis~\cite{liu2019soft} with differentiable rendering applied to reproduce images corresponding to known viewpoints. The 3D meshes are optimized by gradient descent. However, this method requires template meshes for initialization before optimization, which is difficult to obtain due to the complexity of remote sensing scenes. Besides, the images rendered by these methods may suffer from severe artifacts behind occluded areas. Volumetric representation is another approach to representing 3D scenes for novel view synthesis. Early work~\cite{kutulakos2000theory,seitz1999photorealistic} directly represents RGB color information with voxels. Recently, DeepVoxels~\cite{sitzmann2019deepvoxels} proposed a learning-based network to predict 3D feature embedding of each grid in volumetric representation from a set of posed images. Since 3D volume is memory inefficient, the resolution for spatial context needs to be traded off carefully. Although combining with deep convolutional neural networks can compensate for the degradation of rendering high-resolution images from low volume resolutions, the improvement is still limited for remote sensing scenes.

Recent work has demonstrated the capability of implicit neural representation for representing 3D shapes. With implicit neural representations, 3D geometric information can be encoded into the neural network weights by learning the mapping between 3D coordinates and occupancy or signed distance functions~\cite{chen2019learning,genova2020local,park2019deepsdf,mescheder2019occupancy}. By using a MLP model mapping 5D vectors (3D coordinates and 2D view directions) to transparency and color values, NeRF~\cite{mildenhall2020nerf} shows superiority over CNN-based volume rendering methods on view synthesis. Later works like NeRF-W~\cite{martin2021nerf} and NeRF++~\cite{zhang2020nerf++} extend NeRF from object-level to unbounded scenes. For very large-scale scenes, Block-NeRF~\cite{tancik2022block} decomposes the scene into blocks and separately optimizes individual NeRF models. By decoupling the rendering and scene size, Block-NeRF can be scalable to large scenes while allowing individual updates of each block. CityNeRF~\cite{xiangli2021citynerf} achieves city-scale scene reconstruction and proposes a progressive learning method to solve multi-scale problems. Despite the above progress, NeRF-based methods require optimization scene-by-scene, and need sufficient views for supervision. For remote sensing scenes with sparse views, the above conditions cannot be guaranteed, so these methods are difficult to apply. To improve the reconstruction on sparse-view conditions, PixelNeRF~\cite{yu2021pixelnerf} is proposed very recently, which introduces a CNN-based encoder to learn scene prior from one or a few images. PixelNeRF uses a similar idea to the Prior extractor in the proposed method. However, the difference between the proposed method and PixelNeRF is that the former is designed based on implicit multiplane representations while the latter is based on volume rendering and thus suffers from rendering inefficiency.

\subsection{Remote Sensing Image Height/Depth Estimation}

In computer vision and photogrammetry remote sensing, depth/height estimation refers to estimating the distance from an object to the camera with single or multi-view input images. Depth/height estimation and view synthesis are closely related since good estimation results can help to produce good view synthesis. Multi-view stereo based methods have achieved accurate results for remote sensing image depth estimation tasks~\cite{morgan2010precise, mahato2019dense, li2016hierarchical, liu2020dense, facciolo2017automatic, hou2018planarity}. Recently, encoder-decoder based neural networks are introduced to single view height estimation task~\cite{amirkolaee2019height, miclea2021monocular, li2020height, liu2020im2elevation, xing2021gated}. Multi-task learning is also adopted~\cite{carvalho2019multitask, srivastava2017joint} to increase the accuracy of height estimation by jointly learning from semantic labels. However, these methods require ground-truth depth maps or high-resolution digital surface model(DSM) as supervision, which are not always available in practice. Different from the above methods, in this paper, we take advantage of differentiable rendering and self-supervised learning, where we re-project the rendering views to the original view inputs and enforce them to be similar. This way, with the help of multi-view constraints, the depth and 3d structure of the scene can still be understood properly despite the absence of depth ground truth.

% In comparison, we circumvent to introduce 3D supervision in the training stage by differentiable rendering to build loss between the reprojected image and initial image, realizing geometry understanding of scene content.

% \cite{carvalho2019multitask} adopts multi-task learning to increase the accuracy of height estimation by introducing semantic labels. 

%  ******************************Methods****************************** 
\section{Methods}
\label{sec:method}

Given a set of multi-view images of a remote sensing scene as input, our method aims at rendering images corresponding to any new viewpoints. An overview of the proposed method is illustrated in Fig.\ref{fig:overall}. There are two main stages in our method:

\begin{itemize}
\item Stage I: Cross-scene Neural Network Initialization. We train a scene prior extraction network in order to predict object distribution for ImMPI initialization. The heights of common ground objects in remote sensing scenes have potential regularities, which can be applied to narrow down the solution space of scene reconstruction. Taking single image as input, the scene prior extraction network initializes the implicit MPI representation of the input image. For more details, please refer to Section \ref{ssec:initialization}.

\item Stage II: Per-scene Implicit Representation Optimization. After the initialization stage, coarse structure of the scene has been learned in the pre-trained ImMPI model. We then iteratively optimize the parameters of implicit neural network with other viewpoint images. After optimization, accurate novel views can be rendered with the ImMPI model. Details of the per-scene optimization stage can be found in Section \ref{ssec:optimization}.
\end{itemize}

% The initialized ImMPI model has encoded coarse information about the scene.
% ImMPI is a 3D representation of the scene that can be used to synthesize novel view images. 

\subsection{3D Scene Representation}
\label{ssec:3Drepresentation}

We combine implicit neural network with explicit multiplane images to represent remote sensing scenes. Since the optical axis of an onboard camera is almost perpendicular to the ground in remote sensing platforms, the proposed ImMPI is naturally suitable for remote sensing photography. In addition, its efficiency in rendering images via homography warping and differentiable rendering facilitates real-time applications.

\subsubsection{Explicit MPI Representation}
\label{sssec:MPI}

In our method, we use an implicit neural network, i.e., a deep convolutional neural network (CNN) to generate explicit MPI scene representation, where the multi-planary geometry of the scene is encoded in the weights of the CNN. We follow the algorithm~\cite{zhou2018stereo} and divide the 3D space into a collection of RGBA layers $\{(C_{z_{1}}, \sigma_{z_{1}}), (C_{z_{2}}, \sigma_{z_{2}}), \cdots (C_{z_{D}}, \sigma_{z_{D}})\}$ in camera frustum, where $C_{z_{i}}(x, y)$ is a 3-dim vector denoting RGB value at position $[x, y, z_{i}]^{T}$ in camera frustum, $\sigma_{z_{i}}(x, y)$ is a scalar denoting the transmittance of position $[x, y, z_{i}]^{T}$, $D$ is the depth sample number. Let $[x, y]^T$ be a 2D pixel coordinate on the plane, with depth hypothesis $z_{i}$ and pinhole camera intrinsic $K$, we can reconstruct the 3D location $[X, Y, Z]^T$ of the point in Cartesian coordinate as follows:
\begin{equation}
\label{equ:coordinate_transform}
\begin{bmatrix} X\\ Y\\ Z \end{bmatrix} = z_{i} K^{-1}  \begin{bmatrix} x\\ y\\ 1 \end{bmatrix}  = z_{i} \begin{bmatrix} f_{x} & 0 & c_{x}\\ 0 & f_{y} & c_{y}\\ 0 & 0 & 1 \end{bmatrix}^{-1} \begin{bmatrix} x\\ y \\ 1 \end{bmatrix}.
\end{equation}

\subsubsection{Novel View Rendering from MPI}
\label{sssec:MPIrender}

Given a viewpoints denoted as target view, MPI renders the corresponding 2D image as follows. First, map all the planes in MPI to novel view camera frustum by differentiable homography warping. The scene MPI representation is constructed in reference-view camera frustum after initialization, whose intrinsic is noted as $K_{ref}$. Let the novel target-view camera intrinsic be $K_{tgt}$ and the transform matrix between reference-view and target-view be $T_{tgt2ref} = [R_{tgt2ref}, t_{tgt2ref}]$. Then, the correspondence between $[x_{ref}, y_{ref}]$ and $[x_{tgt}, y_{tgt}]$ with respect to pixel coordinate can be calculated as:
\begin{equation}
\label{equ:warping}
\begin{bmatrix}x_{ref}\\ y_{ref}\\ 1\end{bmatrix} = H_{tgt2ref} \begin{bmatrix} x_{tgt}\\ y_{tgt} \\ 1 \end{bmatrix},
\end{equation}
where $H_{tgt2ref}$ denotes the homography warping matrix, calculated by transform matrix $[R_{tgt2ref}, t_{tgt2ref}]$ and depth hypothesis ${z_{i}}$ as follows:
\begin{equation}
H_{tgt2ref} = K_{ref} \left ( R_{tgt2ref} - \frac{t_{tgt2ref} n^T}{z_{i}} \right ) K_{tgt},
\end{equation}
where $n^{T} = [0, 0, 1]^T$ is the normal vector for each plane with respect to reference camera. According to the corresponding relation in Equation \ref{equ:warping}, MPI representation in target-view camera frustum can be thus sampled from the reference-view.

After warping the MPI representation to target camera frustum, we apply differentiable rendering to get novel view 2D images. For each pixel position $(x, y)$ in novel image plane, RGB pixel values are calculated by following the equation below:
\begin{equation}
\label{equ:render}
I = \sum_{i=1}^{N}T_{i}\left (1-exp(-\sigma_{z_{i}}\delta_{z_{i}}) \right )C_{z_{i}}.
\end{equation}
Specifically, $T_{i} = exp(-\sum_{j=1}^{i-1}-\sigma_{z_{j}}\delta_{z_{j}})$ represents the accumulation of transparency from the first plane to the $i$th plane. $\delta_{z_{i}}$ denotes the Euclidean distance between $[x, y, z_{i}]$ and $[x, y, z_{i+1}]$ in Cartesian coordinates, which can be calculated by following the Equation \ref{equ:coordinate_transform}.

Given the MPI representation of the scene and novel viewpoints, the process of rendering at new viewpoints can be finally expressed as:
\begin{equation}
I_{tgt} = \mathcal{R}\left (f_\Phi, T_{tgt2ref}, K_{ref}, K_{tgt} \right ),
\end{equation}
where $\mathcal{R}$ is the MPI renderer defined by Equation \ref{equ:render}. $f_\Phi$ is the multiplane images, where in our method is parameterized by a pre-trained CNN model. Since the MPI rendering is essentially a plane-to-plane warping and ray accumulation process, novel views can be rendered very fast.

As for the depth hypothesis, $z_{i}$ can be estimated from the sparse points by following the structure-from-motion method like COLMAP~\cite{schonberger2016structure} when calculating camera poses. In practice, the depth range $[z_{near}, z_{far}]$ of remote sensing scenes can be much larger. Therefore, the depth sampling strategy of MPI is crucial. With a pre-defined depth sample number $D$, we evenly sample hypotheses on reciprocal depth space by following the strategy~\cite{xu2020learning}:
\begin{equation}
\frac{1}{z_{i}} = \frac{1}{z_{far}} +  \frac{i-1}{D} \left ( \frac{1}{z_{near}} - \frac{1}{z_{far}}\right ).
\end{equation}
The above operation helps to make the rendering process applicable to complex and large depth range remote sensing scenes.
Finally, the depth map $I_{depth}$ under the query view can also be calculated in a similar way as the rendered image:
\begin{equation}
\label{equ:depth}
I_{depth} = \sum_{i=1}^{N}T_{i}(1-exp(-\sigma_{z_{i}}\delta_{z_{i}}))z_{i}.
\end{equation}

% MPI renders novel views very fast due to the multi-plane representation of the 3D space. 

% generated by scene depth range $[z_{near}, z_{far}]$,

\begin{figure}[t]
    \centering
    \includegraphics[width=0.47\textwidth]{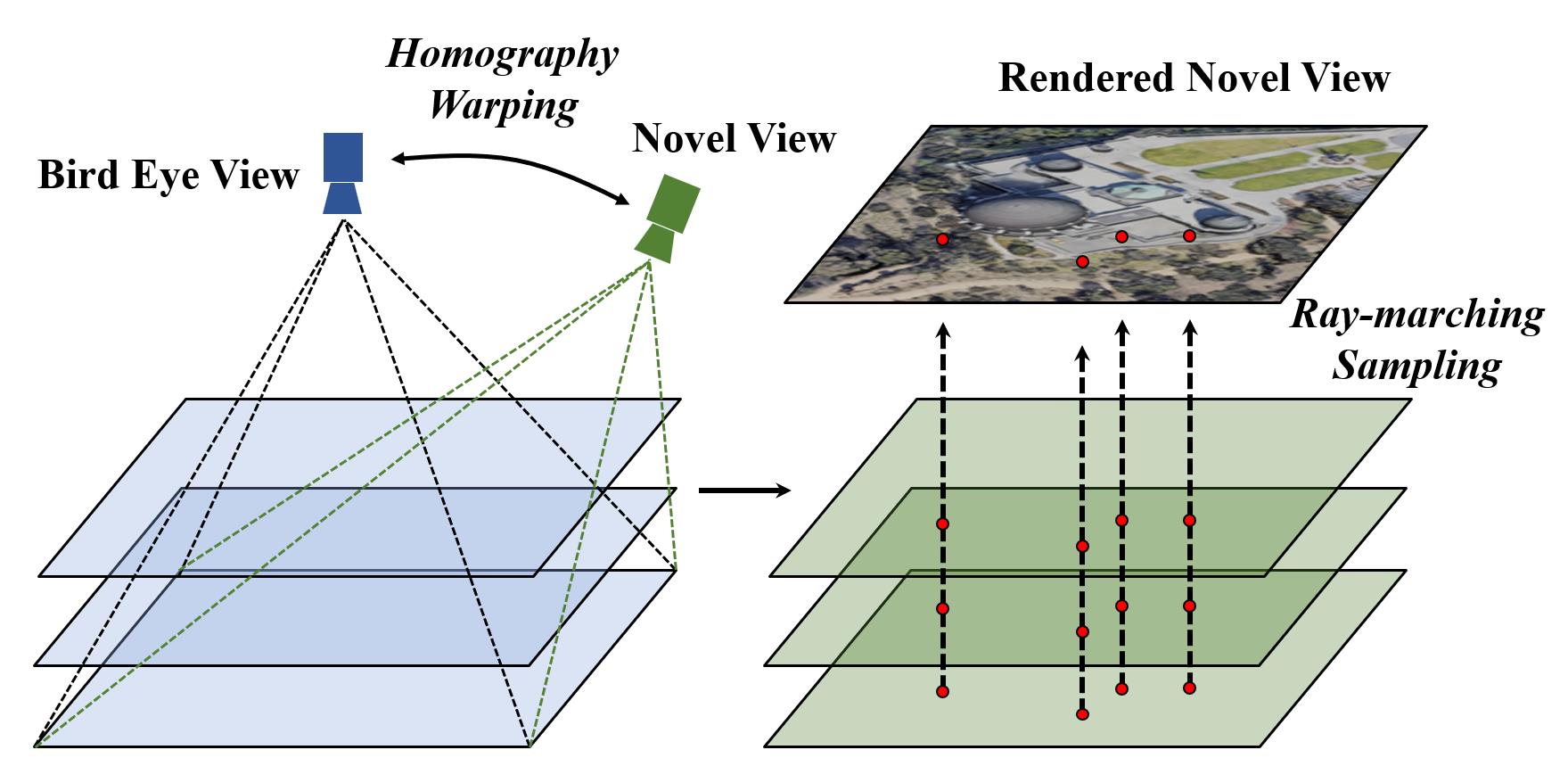}
    \caption{Illustration of the view rendering process from MPI. Reference-view and target-view are marked with blue and green cameras. The MPI representation is constructed with respect to reference-view frustum initially. To render novel views, the MPI is firstly transformed to the target-view coordinate by using Homography warping. We then apply ray-marching sampling to render the image.}
    \label{fig:flowchart_mpirendering}
\end{figure}

\subsection{Cross-scene Neural Network Initialization}
\label{ssec:initialization}

In this subsection, we introduce the Cross-scene Neural Network Initialization method. We introduce a scene prior extraction network and present a self-supervised learning method to learn 3D distribution priors from a single image. The key is the use of differentiable rendering and enforcing the projected views to be similar to the source views. The MPI can be thus roughly estimated with the above view constraints. The overview process is shown in Fig.~\ref{fig:overall}.

% and train it on remote sensing multi-view depth estimation dataset, which supplies grouped posed images.

% We present a learning based network to extract remote sensing object distribution priors from a single image and train it on remote sensing multi-view depth estimation dataset, which supplies grouped posed images.

\subsubsection{Network Design and Training Process}
\label{sssec:network}
\begin{figure}[t]
    \centering
    \includegraphics[width=0.5\textwidth]{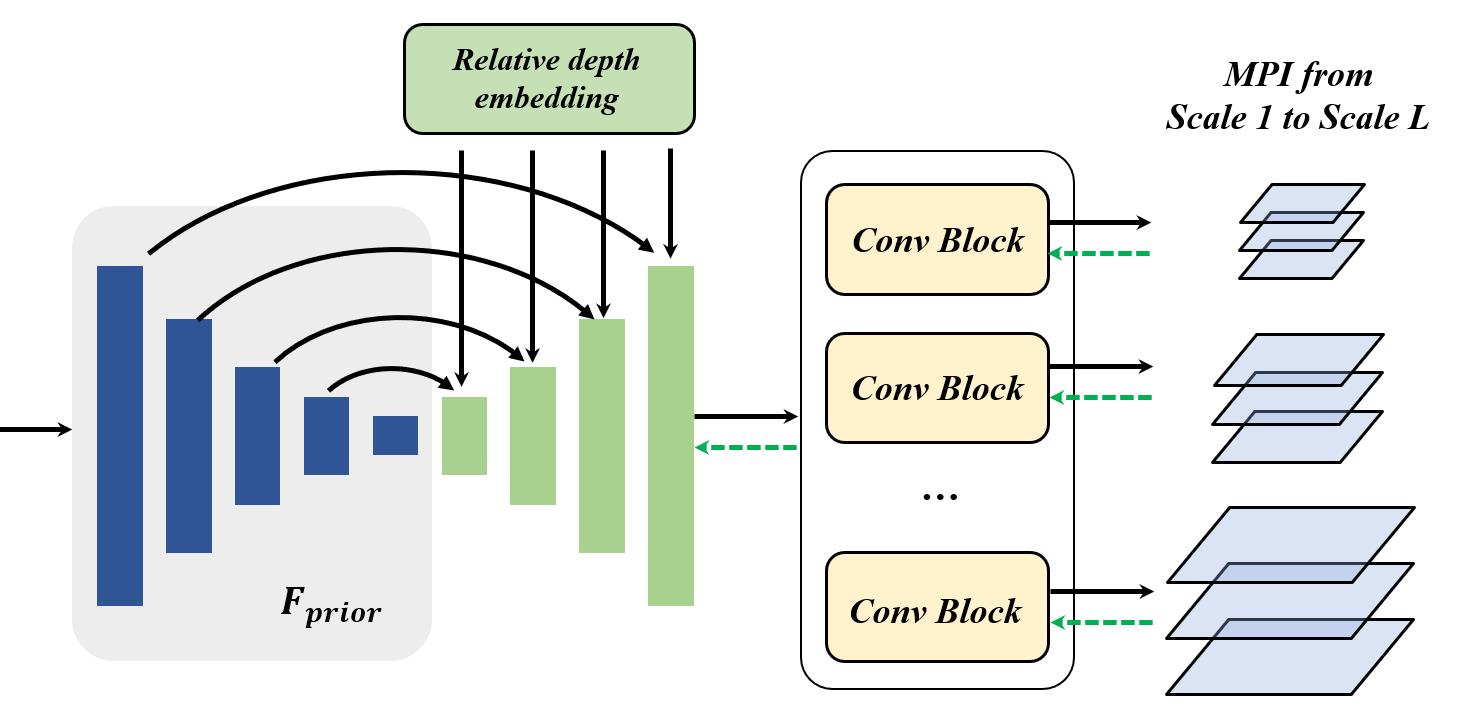}
    \caption{Architecture of the MPI generator. Depth embedding vectors are concatenated with five-scale $F_{prior}$. Each scale feature passes through the convBlock to generate corresponding explicit MPI.}
    \label{fig:mpigenerator}
\end{figure}

Given an image denoted as the reference view, the prior extractor encodes the 2D image feature as $F_{prior}$. Then, the feature is input to the MPI Generator to obtain the initial MPI representation of the scene. Details of the network design are illustrated in Fig.~\ref{fig:mpigenerator}. Specifically, we adopt ResNet18~\cite{he2016deep} as the backbone of our Prior extractor. The Prior extractor takes in a single image and produces multi-scale features. Then, the MPI generator takes in multi-scale features and produces multi-scale MPI representations. We set the number of features scales to 5 and the number of MPI scales to 4 in our method.

%  corresponding to x1, x2, x4, x8 downsample rate respectively.

\textbf{Relative Depth Embedding}. Remote sensing scenes may have very different depth ranges. However, we expect to train a generic model that captures depth priors as comprehensively as possible for MPI initialization. Considering that it's difficult to recover the accurate absolute depth from a single image directly, we adopt relative depth as the depth value $z_{i}$ of different planes in MPI. Specifically, suppose the depth sample number is $D$, we apply a 1-dim positional embedding to $d_{i} = \left\{0, 1, 2, ..., D-1 \right\}$ by following the Equation \ref{equ:depth_embed} presented in \cite{tancik2020fourier}:
\begin{equation}
\label{equ:depth_embed}
\begin{split}
\mathcal{\gamma}(d_{i}) = [\sin(2^{0}\pi d_{i}), \cos(2^{0}\pi d_{i}), ..., \\
\sin(2^{L-1}\pi d_{i}), \cos(2^{L-1}\pi d_{i})].
\end{split}
\end{equation}
Then the depth embedding vectors are merged with multi-scale features $F_{prior}$. 
When rendering the target view image, we map the $d_{i}$ of MPI plane to the absolute depth values $z_{i}$ with respect to the depth range of reference image.

\textbf{Training Process}. The scene prior extraction network is trained by using a large number of images from different scenes. The network can be trained simply on any remote sensing multi-view stereo datasets. Taking multi-scale MPI representation generated from the corresponding reference image as input, we can render target-view images according to the procedure illustrated in \ref{ssec:3Drepresentation}.

\subsubsection{Loss functions for Cross-scene Training}
\label{sssec:loss_initialization}

We train our scene prior extraction network on the WHU MVS/Stereo Dataset~\cite{liu2020novel}. This dataset provides four posed neighbor images for each reference image. We train the model with reference-target paired images. Note that only 2D RGB images in the WHU MVS/Stereo dataset are used for training, and no 3D supervision such as depth map is introduced. To optimize the Prior extractor and MPI generator, we minimize the L1 loss $\mathcal{L}_{L1}$ and Structure Similarity Index Measure (SSIM) loss~\cite{wang2004image} $\mathcal{L}_{ssim}$ between the rendered image and corresponding ground-truth:
\begin{equation}
\mathcal{L}_{prior}(\theta,\phi) = \sum_{s=1}^{L} \lambda_1\mathcal{L}_{L1}(I_{s}^{tgt}, I_{s}^{gt}) + \lambda_2\mathcal{L}_{ssim}(I_{s}^{tgt}, I_{s}^{gt}),
\end{equation}
where $\theta$, $\phi$ are the network parameters of the Prior extractor and MPI generator. $s$ refers to the scale of MPI and $L$ is the total number of scales. $\lambda_{L1}$ and $\lambda_{ssim}$ are pre-defined weights to balance the two loss terms. We set $\lambda_1=2.0$ and $\lambda_2=1.0$  respectively. Since the networks and the MPI rendering process are all differentiable, the networks can be trained in an end-to-end fashion with the above losses.

% \begin{equation}
% \theta^{*},\phi^{*} = \mathop{\arg\min}\limits_{\theta,\;\phi} \sum_{s=1}^{4} \mathcal{L}_{prior}\left (I_{s}^{tgt}, \; I_{s}^{gt} \right )
% \end{equation}

% network parameters $\theta$, $\phi$:

%  The loss function is defined as following. The hyperparameters $\lambda_{L1}$, $\lambda_{ssim}$ equal to 2.0, 1.0 respectively.
% \begin{equation}
% \mathcal{L}_{prior} = \lambda_{L1}\mathcal{L}_{L1} + \lambda_{ssim}\mathcal{L}_{ssim}
% \end{equation}

% Multi view stereo dataset provides posed images, which can be utilized for Across Scene Initialization neural network training to extract prior features. 

\begin{algorithm}[t!]
\caption{Novel View Synthesis with the proposed ImMPI model.} 
\label{alg-inference} 
\KwIn{$\mathbf{B}=\{(\mathbf{I}^{src}_{n},\mathbf{T}^{src}_{n}, \mathbf{K}^{src}_{n}) | n = 1:N \}$ (images and camera parameters of training views)}
\KwIn{$(\mathbf{T}^{tgt}, \mathbf{K}^{tgt})$ (novel viewpoints)}
\KwIn{$\mathbf{Iteration}$ (iterate number during optimization)}
\KwOut{$\mathbf{ImMPI}$ (scene representation)}
\KwOut{$\mathbf{I}^{tgt}$ (novel view RGB image)}
\BlankLine
    // step1: extract priors and initialize ImMPI \\
        \quad // select a training view denoted as reference view \\
        \quad $(\mathbf{I}^{ref}, \mathbf{T}^{ref}, \mathbf{K}^{ref}) = Sample(\mathbf{B})$  \\
        \quad // extract prior with pretrained Prior extractor $f_{\theta}$ \\
        \quad $\mathbf{F}_{prior} = f_{\theta^{\star}}(\mathbf{I}^{ref})$ \\
    // step2: optimize ImMPI with training views \\
    \For{$i$ in $1:\mathbf{Iteration}$}
    {
        \For{($\mathbf{I}^{src}_{n}, \mathbf{T}^{src}_{n}, \mathbf{K}^{src}_{n}$) in $\mathbf{B}$}
        {
            // render images \\
            $\mathbf{MPI} = f_{\phi}(\mathbf{F}_{prior})$ \\
            $\mathbf{I}^{syn} = \mathcal{R}(\mathbf{MPI}, \mathbf{T}^{src}_{n}, \mathbf{K}^{src}_{n}, \mathbf{K}^{ref})$ \\
            // gradient descent to optimize ImMPI  \\
            % $\phi$ \xleftarrow{$GradDescent(\mathbf{I}^{syn}_{n}, \mathbf{I}^{src}_{n})$} \\
        }
    }
    \quad // $\textbf{ImMPI}^{\star}$: scene info encoded in $F_{prior}$ and $\phi^{\star}$ \\
    \quad $\textbf{MPI}^{\star} = f_{\phi^{\star}}(f_{\theta^{\star}}(\mathbf{F}_{prior}))$ \\
    // step3: render novel view \\
    \quad $\mathbf{I}^{tgt} = \mathcal{R}(\mathbf{MPI}^{\star}, \mathbf{T}^{tgt}, \mathbf{K}^{tgt}, \mathbf{K}^{ref})$ \\
\end{algorithm}

\subsection{Per-scene Implicit Representation Optimization}
\label{ssec:optimization}

Since the initialization stage only utilizes information from one perspective, the initialized MPI representation is not accurate enough and may produce artifacts in occluded areas. In the optimization stage, we further optimize the implicit MPI representation of a specific scene iteratively with images from other viewpoints.

\subsubsection{Optimization}
\label{sssec:optimization_process} 
The per-scene optimization process is illustrated in Fig.~\ref{fig:overall}. With scene prior $F_{prior}$ learned from the reference image, we can generate an initial implicit MPI representation. During the optimization process, the parameters of the Prior extractor are fixed and the weights of the MPI generator network are updated. We do not directly optimize the MPI pixels for two reasons. On the one hand, it's very easy to overfit on training views because of too many parameters of explicit MPI. On the other hand, when directly optimizing the RGBA values of MPI, each position is treated individually ignoring the local similarity. In comparison, optimizing the parameters of the CNN-based MPI generator preserves spatial continuity, which in turn can bring smoothness to the rendered image.

\subsubsection{Loss Functions for Optimization}
\label{sssec:loss_optimization} 

Unlike NeRF~\cite{mildenhall2020nerf} that randomly samples points in training images, the proposed ImMPI renders the entire image of training views. The losses can therefore be applied at the image level. Given multi-scale MPIs and transformation matrix of training view, multi-scale images $\{I_{s} | s=1,...,L\}$ can be rendered according to Equation~\ref{equ:render}. Following the objective function designed in~\cite{zhou2018stereo}, the loss function is calculated between rendered and ground-truth images. The loss function is defined as:
\begin{equation}
\begin{split}
\mathcal{L}_{opt}(\phi) &= \sum_{s=1}^{L} \sum_{i=1}^{N}\beta_1\mathcal{L}_{L1}(I_{s, i}^{tgt}, \; I_{s, i}^{gt})\\
&+ \beta_2\mathcal{L}_{ssim}(I_{s, i}^{tgt}, \; I_{s, i}^{gt}) + \beta_3\mathcal{L}_{lpips}(I_{s, i}^{tgt}, \; I_{s, i}^{gt})
\end{split}
\end{equation}
where $N$ is the number of target views and $L$ is the number of scales. $\mathcal{L}_{L1}$, $\mathcal{L}_{ssim}$, and $\mathcal{L}_{lpips}$ represent pixel-wise L1 loss, SSIM loss, and the Learned Perceptual Image Patch Similarity (LPIPS) loss~\cite{zhang2018unreasonable}, respectively. The LPIPS loss is computed as the distance between two images on their multi-scale features produced by the VGG-19 networks~\cite{simonyan2014very}, which aims at improving the visual fidelity of the rendering outputs. $\beta_1$, $\beta_2$, and $\beta_3$ are their pre-defined balancing weights. 

Finally, at novel view rendering phase, given a novel viewpoint, the rendering details of our method are shown in Algorithm \ref{alg-inference}.

% There are two terms in RGB loss function: L1 loss $\mathcal{L}_{L1}$ and SSIM loss $\mathcal{L}_{ssim}$. Besides, we introduce the perceptual loss $\mathcal{L}_{lpips}$ to optimization process, which evaluates the similarity between multi-scale features produced by VGG-19.
% \begin{equation}
% \mathcal{L}_{optim} = \lambda_{L1}\mathcal{L}_{L1} + \lambda_{ssim}\mathcal{L}_{ssim} + \lambda_{lpips}\mathcal{L}_{lpips}
% \end{equation}

% \begin{equation}
% \phi^{*} = \mathop{\arg\min}\limits_{\phi} \sum_{s=1}^{4} \sum_{i=1}^{N}\mathcal{L}_{optim}\left (I_{s, i}^{tgt}, \; I_{s, i}^{gt} \right )
% \end{equation}

%  ******************************Experiments****************************** 
\section{Experiments}

In this section, we first introduce the dataset used for training cross-scene initialization and our new dataset for per-scene optimization. Then, experiments are conducted on our new dataset and compare with other view synthesis methods. Finally, controlled experiments and ablation analysis are given to verify the effectiveness of our method.

% Our proposed method consists of two parts: Across Scene Neural Network Initialization and Per Scene Optimization. The former is learning-based and the latter is based on iterative optimization. We first adopt the WHU MVS/Stereo Dataset proposed in \cite{liu2020novel} to train Across Scene Neural Network Initialization. As for Per Scene Optimization, we present a new dataset for novel view synthesis in Remote Sensing scenes termed as LEVIR-NVS. We verify the effectiveness of our method on this dataset.

\subsection{Experimental Setup}
\label{ssec:setup}

\textbf{WHU MVS/Stereo Dataset}\cite{liu2020novel} is a public large-scale Earth surface reconstruction dataset. It consists of 1776 images captured in 11 strips by UAV. The covered area contained dense and tall buildings, sparse factories, mountains covered with forests, and some bare ground and rivers. This dataset is mainly used for multi-view depth estimation tasks. We train the cross-scene initialization based on this dataset.

\textbf{LEVIR-NVS Dataset} is our newly proposed dataset for remote sensing image novel view synthesis\footnote{LEVIR is the laboratory's name where the authors of this paper are in. NVS is short for ``Novel View Synthesis''}. We use Blender to acquire multi-view 2D images of 3D scene models captured from Google Earth. The dataset consists of 16 scenes, including mountains, cities, villages, buildings, etc. Each scene has 21 multi-view images of size 512x512, 11 views are used for training and the rest are used for testing. Pose transformations such as wrapping and swinging in actual aerial photography are included during the simulation process.

\textbf{Implementation details}. Our model is implemented on Pytorch and is trained using a single GeForce RTX 3090 GPU. For the cross-scene initialization training, we apply Adam Optimizer\cite{kingma2014adam} to optimize the model. The learning rate is set to 0.0001 initially and decays 0.5 times per 40 epochs during 200 training epochs. During the per scene optimization, we also apply the Adam optimizer and the learning rate is set to 0.001. For each scene in LEVIR-NVS, the optimization can converge in less than 500 iterations.

\textbf{Evaluation Metrics}. Similar to the metrics adopted in previous novel view synthesis literature~\cite{mildenhall2020nerf, zhang2020nerf++, yu2021pixelnerf, li2021mine}, we apply PSNR, SSIM, LPIPS~\cite{zhang2018unreasonable} to evaluate the rendering accuracy. PSNR and SSIM evaluate pixel-level differences between rendered images and ground-truth images, and LPIPS utilizes a VGG network to evaluate image similarity at feature level.

\subsection{Comparison to other methods}
\label{ssec:comparison}

We compare our method with two state-of-the-art novel view synthesis methods: NeRF~\cite{mildenhall2020nerf}, NeRF++~\cite{zhang2020nerf++}.

\begin{table*}[]
    \centering
    \caption{Quantitative comparison of Train-view/Test-view with different Novel View Synthesis methods on LEVIR-NVS dataset.}
    \resizebox{\textwidth}{!}{
    \begin{tabular}{c|ccc|ccc|ccc}
    \toprule
    \multicolumn{1}{c|}{} & \multicolumn{3}{c|}{\textbf{PSNR}$\uparrow$} & \multicolumn{3}{c|}{\textbf{SSIM}$\uparrow$} & \multicolumn{3}{c}{\textbf{LPIPS}$\downarrow$} \\
    \textbf{Scenes} & NeRF~\cite{mildenhall2020nerf} & NeRF++~\cite{zhang2020nerf++} & ImMPI (Ours) & NeRF~\cite{mildenhall2020nerf} & NeRF++~\cite{zhang2020nerf++} & ImMPI (Ours) & NeRF~\cite{mildenhall2020nerf} & NeRF++~\cite{zhang2020nerf++} & ImMPI (Ours) \\
    \midrule
    Building\#1             & 20.76 / 12.21 & 22.76 / 14.19 & 24.92 / 24.77          & 0.533 / 0.147 & 0.649 / 0.280 & 0.867 / 0.865          & 0.530 / 0.652 & 0.434 / 0.601 & 0.150 / 0.151          \\
    Building\#2             & 20.34 / 12.38 & 24.72 / 18.48 & 23.31 / 22.73          & 0.444 / 0.184 & 0.711 / 0.483 & 0.783 / 0.776          & 0.573 / 0.668 & 0.384 / 0.507 & 0.217 / 0.218          \\
    College                 & 23.65 / 12.16 & 26.66 / 18.57 & 26.17 / 25.71          & 0.591 / 0.181 & 0.746 / 0.471 & 0.820 / 0.817          & 0.475 / 0.645 & 0.319 / 0.474 & 0.201 / 0.203          \\
    Mountain\#1             & 25.17 / 16.29 & 26.40 / 23.65 & 30.23 / 29.88          & 0.565 / 0.290 & 0.616 / 0.563 & 0.854 / 0.854          & 0.556 / 0.674 & 0.484 / 0.508 & 0.187 / 0.185          \\
    Mountian\#2             & 24.84 / 12.48 & 26.08 / 24.45 & 29.56 / 29.37          & 0.482 / 0.168 & 0.563 / 0.519 & 0.844 / 0.843          & 0.551 / 0.631 & 0.481 / 0.501 & 0.172 / 0.173          \\
    Mountain\#3             & 27.65 / 18.86 & 30.43 / 24.14 & 33.02 / 32.81          & 0.603 / 0.395 & 0.737 / 0.541 & 0.880 / 0.878          & 0.506 / 0.602 & 0.365 / 0.483 & 0.156 / 0.157          \\
    Observation             & 21.42 / 12.48 & 24.71 / 16.98 & 23.04 / 22.54          & 0.498 / 0.168 & 0.701 / 0.386 & 0.728 / 0.718          & 0.501 / 0.631 & 0.353 / 0.505 & 0.267 / 0.272          \\
    Church                  & 18.75 / 10.59 & 23.70 / 13.50 & 21.60 / 21.04          & 0.421 / 0.126 & 0.701 / 0.302 & 0.729 / 0.720          & 0.563 / 0.654 & 0.374 / 0.564 & 0.254 / 0.258          \\
    Town\#1                 & 21.05 / 12.79 & 25.89 / 18.26 & 26.34 / 25.88          & 0.451 / 0.191 & 0.747 / 0.401 & 0.849 / 0.844          & 0.553 / 0.631 & 0.328 / 0.510 & 0.163 / 0.167          \\
    Town\#2                 & 19.62 / 11.43 & 25.22 / 17.03 & 25.89 / 25.31          & 0.434 / 0.187 & 0.738 / 0.417 & 0.855 / 0.850          & 0.573 / 0.654 & 0.336 / 0.500 & 0.156 / 0.158          \\
    Town\#3                 & 20.44 / 12.25 & 26.52 / 18.30 & 26.23 / 25.68          & 0.470 / 0.230 & 0.786 / 0.546 & 0.840 / 0.834          & 0.561 / 0.658 & 0.307 / 0.453 & 0.187 / 0.190          \\
    Stadium                 & 21.40 / 14.11 & 26.64 / 19.26 & 26.69 / 26.50          & 0.497 / 0.286 & 0.753 / 0.504 & 0.878 / 0.876          & 0.540 / 0.649 & 0.313 / 0.466 & 0.123 / 0.125          \\
    Factory                 & 20.26 / 11.79 & 25.51 / 13.13 & 28.15 / 28.08          & 0.495 / 0.242 & 0.749 / 0.296 & 0.908 / 0.907          & 0.572 / 0.656 & 0.363 / 0.597 & 0.109 / 0.109          \\
    Park                    & 20.04 / 13.86 & 25.41 / 17.70 & 27.87 / 27.81          & 0.427 / 0.228 & 0.736 / 0.407 & 0.896 / 0.896          & 0.576 / 0.653 & 0.362 / 0.528 & 0.123 / 0.124          \\
    School                  & 20.94 / 12.33 & 24.90 / 19.50 & 25.74 / 25.33          & 0.503 / 0.268 & 0.673 / 0.516 & 0.830 / 0.825          & 0.559 / 0.669 & 0.399 / 0.479 & 0.163 / 0.165          \\
    Downtown                & 19.26 / 11.62 & 24.69 / 15.17 & 24.99 / 24.24          & 0.427 / 0.163 & 0.718 / 0.331 & 0.825 / 0.816          & 0.585 / 0.665 & 0.393 / 0.578 & 0.201 / 0.205          \\
    \midrule
    mean                    & 21.02 / 12.68 & 25.59 / 17.96 & \textbf{26.34 / 25.95} & 0.481 / 0.201 & 0.705 / 0.425 & \textbf{0.835 / 0.831} & 0.546 / 0.647 & 0.371 / 0.514 & \textbf{0.172 / 0.173} \\
    \bottomrule
    \end{tabular}
    }
    \label{tab:comparison_train/test}
\end{table*}

\begin{itemize}
    \item NeRF~\cite{mildenhall2020nerf} is a neural implicit representation based method for novel view synthesis. NeRF represents the scene volume with a MLP model mapping 5D vectors (3D coordinates and 2D view directions) to transparency and color values. NeRF optimizes the MLP representation by a set of posed images during training. The optimized MLP then can be used to render novel views with conventional volume rendering approaches. NeRF assumes the entire scene to be contained in a bounded volume and the training views are captured from 360-degree viewpoints distributed on a hemisphere. This assumption makes NeRF hard to be used in large-scale scenes.
    % with sparse perspectives of unbounded scene, NeRF may fail to generalize to novel test views cause the precise scene geometry is hard to reconstructed.
    
    \item NeRF++~\cite{zhang2020nerf++} propose to apply NeRF to 360 degree captures of objects within large-scale, unbounded  scenes. The authors proposed a novel spatial parameterization scheme called inverted sphere parameterization as a remedy for vanilla NeRF. Specifically, NeRF++ models the scene space with two separate NeRFs, an inner unit sphere and an outer volume, representing foreground and background respectively. After optimizing the models individually, the render results are composited together to generate final novel view images.
    % Because of the special parameterization mechanism of the algorithm, scene need to be normalized by putting all the camera centers inside the unit sphere, which is benefit for large scale scene reconstruction but leads to lost of real scale.
\end{itemize}

\begin{figure*}
    \centering
    \includegraphics[width=\textwidth]{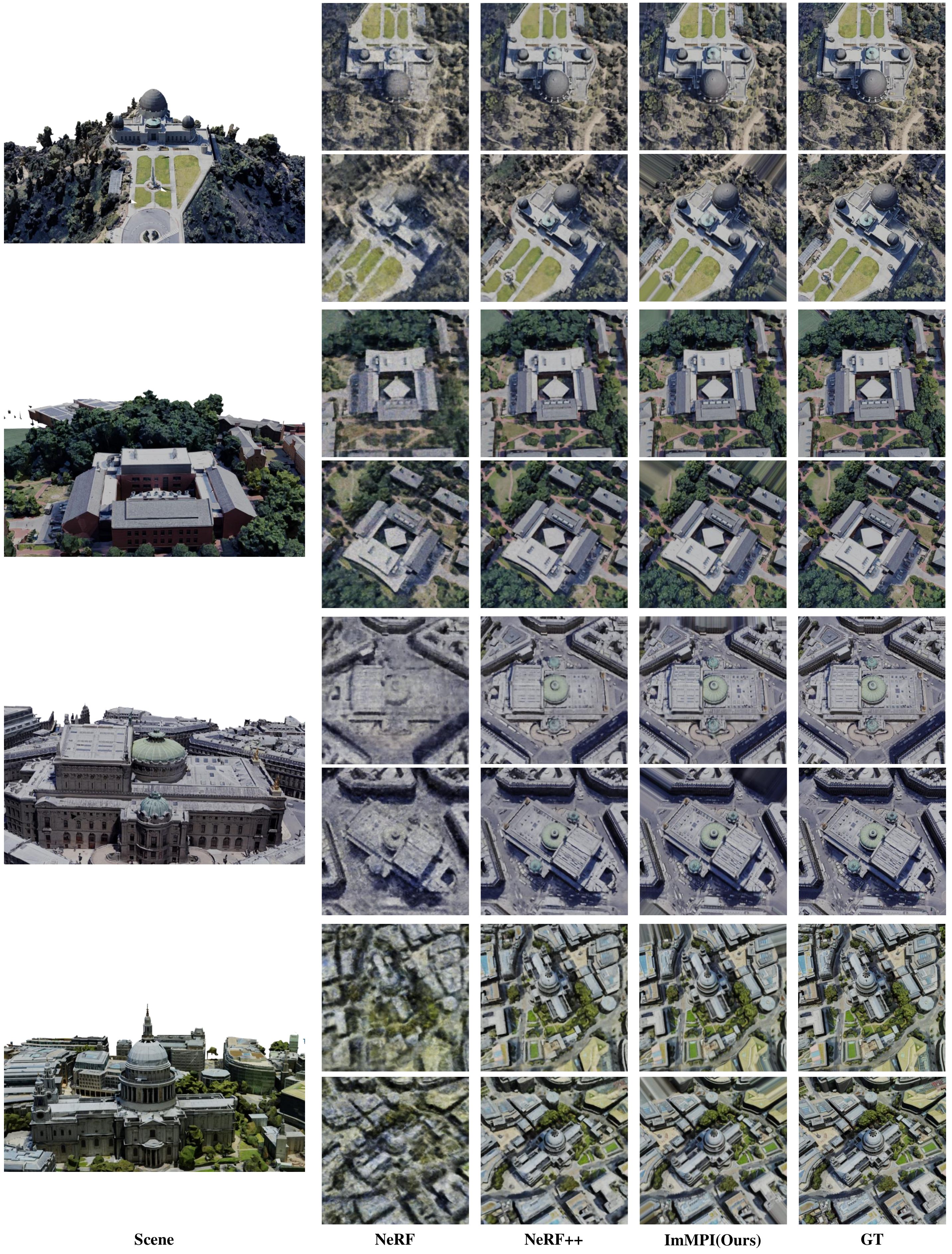}
    \caption{Qualitative comparison of view rendering with training camera poses on LEVIR-NVS dataset.}
    \label{fig:visualization_train}
\end{figure*}

\begin{figure*}
    \centering
    \includegraphics[width=\textwidth]{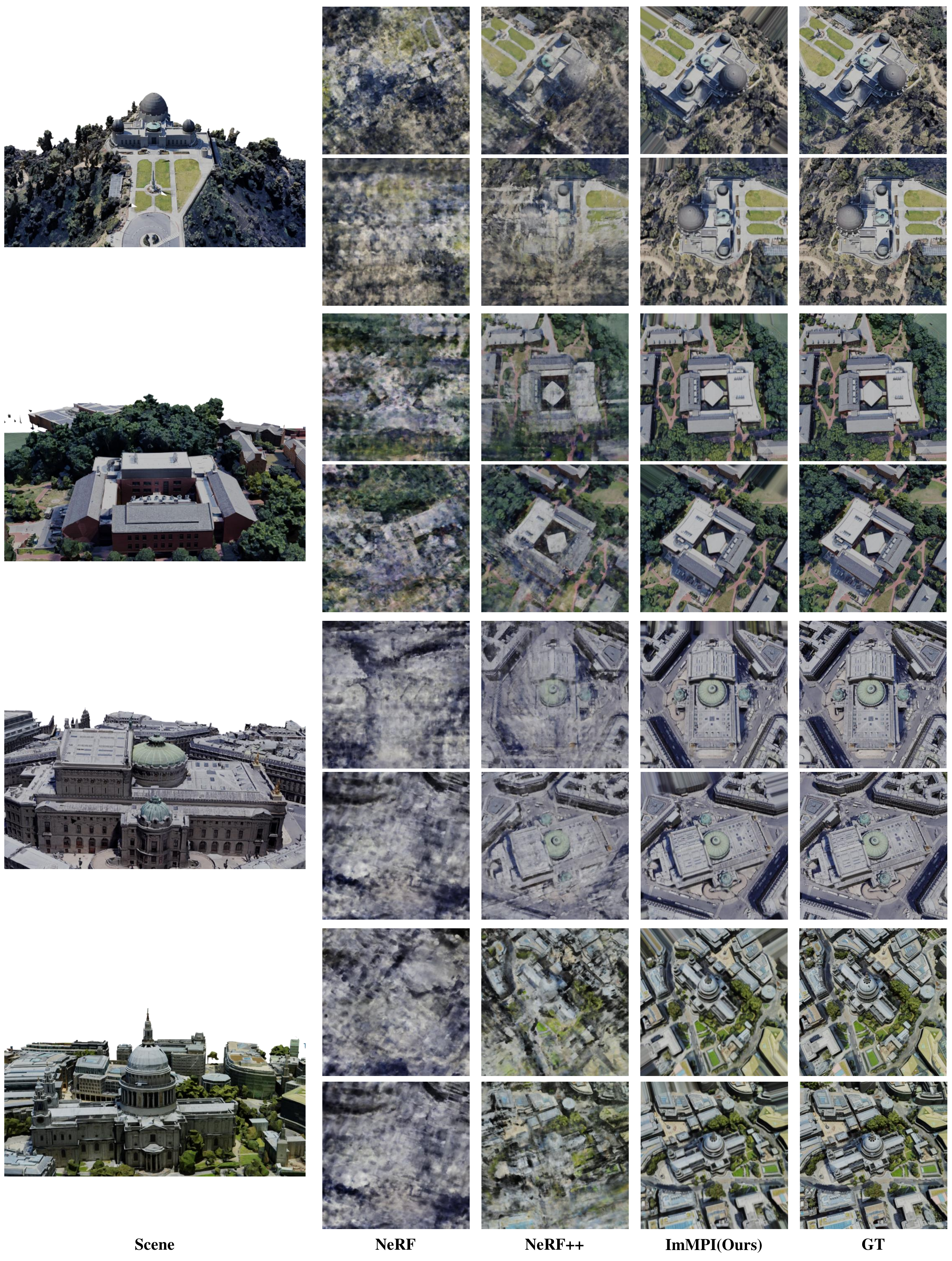}
    \caption{Qualitative comparison of test view rendering (novel view synthesis) on LEVIR-NVS dataset. The views in this figure have not been seen in the pre-training and optimization process.}
    \label{fig:visualization_test}
\end{figure*}

Fig.~\ref{fig:visualization_train} and Fig.~\ref{fig:visualization_test} show qualitative comparison between our method and the two comparison methods. We can see that our method brings a significant increase in rendering fidelity. Although NeRF and NeRF++ can achieve decent results in the training views, the rendering results on test views are not satisfactory. Both NeRF and NeRF++ suffer from noticeable blurring, loss of details and severe artifacts on test viewpoints. We attribute this phenomenon to the fact that the sparse views of remote sensing scenes do not provide sufficient supervision for these methods to achieve accurate reconstruction. Benefiting from the efficient ImMPI initialization, our method does not overfit the training viewpoint and renders realistic images corresponding to given test viewpoints. The edges and textures of objects in the scene are clear and highly similar to the ground-truth. 

Quantitative evaluations of different methods are given in Tab.~\ref{tab:comparison_train/test}. In Train-view, NeRF and NeRF++ can produce relatively high quantitative accuracy in terms of PSNR, SSIM and LPIPS. However, we see a significant drop in the accuracy of test views. Overfitting to training viewpoints leads to inaccurate scene reconstruction, which in turn affects the synthesis results of new viewpoints. As a comparison, our method performs much better than other algorithms in test views, which suggests that ImMPI achieves more accurate reconstruction of remote sensing scenes and can produce high-quality novel view images.

\subsection{Efficiency in Optimization and Rendering}
\label{ssec:model_efficiency}

In this section, we analyze the efficiency of our method during Per Scene Optimization and rendering. Tab.~\ref{tab:time_comparison} shows the time consumption of different methods. In this table, ``Pre-Training'' refers to the pre-training of the methods. ``Optimization'' denotes the per-scene optimization process. ``Rendering'' means the time consumption of synthesizing novel views. We can see that NeRF and NeRF++ are purely optimization-based methods without pre-training, and the proposed ImMPI needs to be pre-trained with extra 21h beforehand. At the optimization phase, both NeRF and NeRF++ take more than an hour to optimize on each scene while the proposed ImMPI only takes 30 minutes. At the rendering phase, our method only takes only 1 seconds to render per frame from the optimized 3D representation, which is at least 20 times faster than other methods.

NeRF-based methods draw the image pixel-by-pixel in ray-tracing manner. Each pixel in novel view needs to sample points along the ray to calculate the RGB value following the Equation~\ref{equ:render}. On the one hand, every point on the ray needs the MLP network forward once to obtain the final value $C_{z}, \sigma_{z}$ requiring extensive computation. Purely implicit 3D scene representation and unaccelerated ray-tracing rendering lead to inefficiencies in the optimization process and image rendering. On the other hand, the weights of the MLP network in NeRF and NeRF++ are randomly initialized resulting in large solution space during optimization, which not only leads to time consumption, but also is prone to degenerate geometry estimation of the input scene.

We attribute the faster convergence speed of ImMPI during per-scene optimization to the following reasons: a) ImMPI generates explicit MPI representations whose high efficiency speeds up optimization and rendering process. b) With the learning-based initialization, cross-scene priors are encoded in the ImMPI model, and thus fewer iterations are required in the optimization stage. c) our method inherits the advantages of implicit representation encoding the scene information in network weights, with less optimal variables than explicit representations.

\begin{table}[]
    \centering
    \caption{Quantitative comparison of optimization and rendering speed between different methods. 
    % Hyperparameters need to be tuned individually and time consumption varies slightly across different scenes.
    }
    \resizebox{0.5\textwidth}{!}{
    \begin{tabular}{c|c|ccc}
    \toprule
    Method & imageSize & Pre-Train & Optimization     & Rendering \\
    \midrule
    NeRF~\cite{mildenhall2020nerf}    & 512x512     & -            & \textgreater{}90min & \textgreater{}20s       \\
    NeRF++~\cite{zhang2020nerf++}  & 512x512     & -            & \textgreater{}60min & \textgreater{}20s       \\
    ImMPI (ours)  & 512x512     & 21h          & \textless{}30min & \textless{}1s       \\
    \bottomrule
    \end{tabular}
    }
    \label{tab:time_comparison}
\end{table}

\begin{figure}[]
    \centering
    \includegraphics[width=0.5\textwidth]{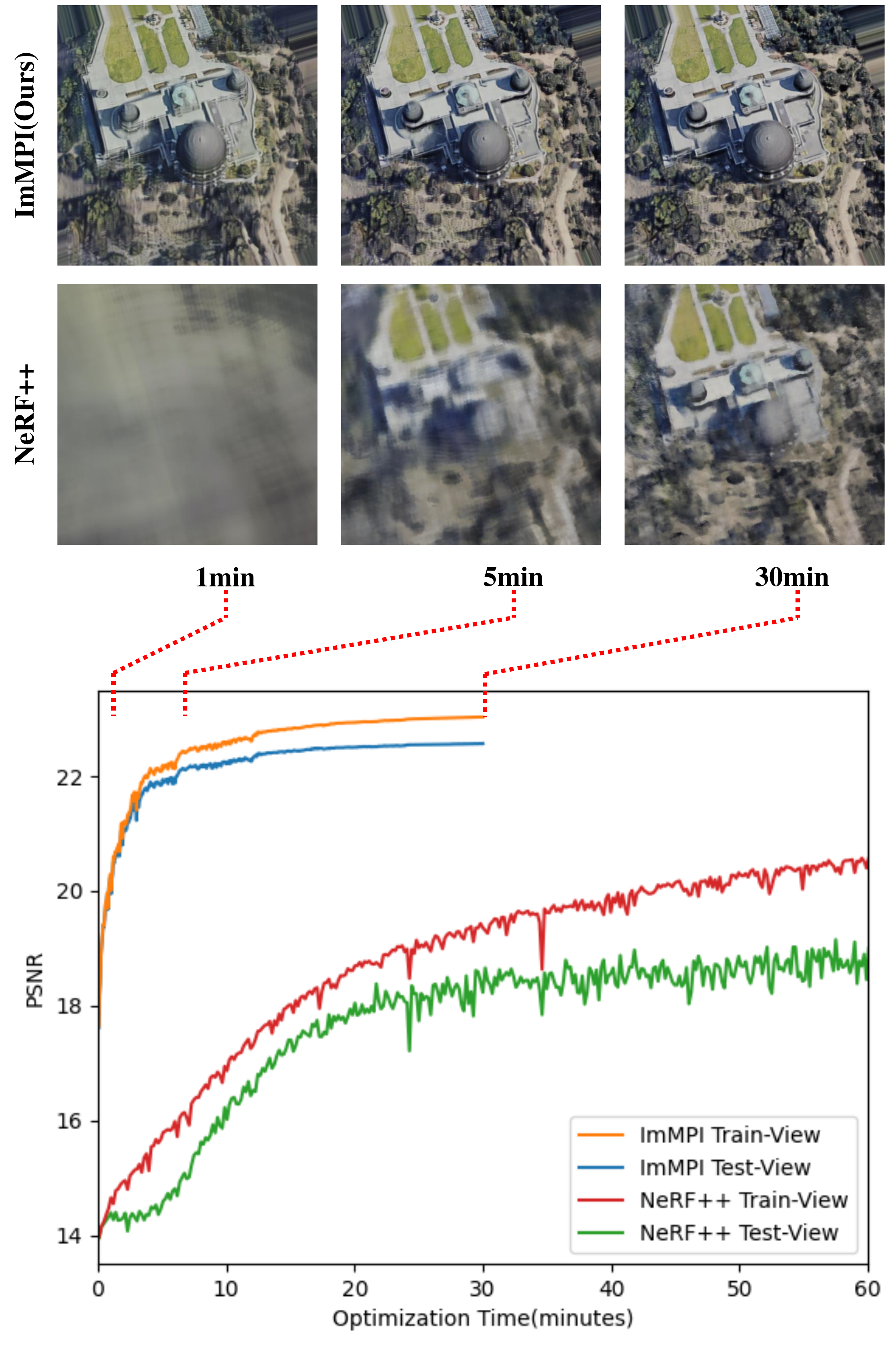}
    \caption{Optimization comparison between the proposed ImMPI and NeRF++~\cite{zhang2020nerf++}. ImMPI is 2 times faster than NeRF++~\cite{zhang2020nerf++} and performs better in test view rendering quality while NeRF++~\cite{zhang2020nerf++} is prone to overfit with more iterations due to sparse view inputs.}
    \label{fig:visualization_optimize}
\end{figure}

\begin{figure*}[t]
    \centering
    \includegraphics[width=\textwidth]{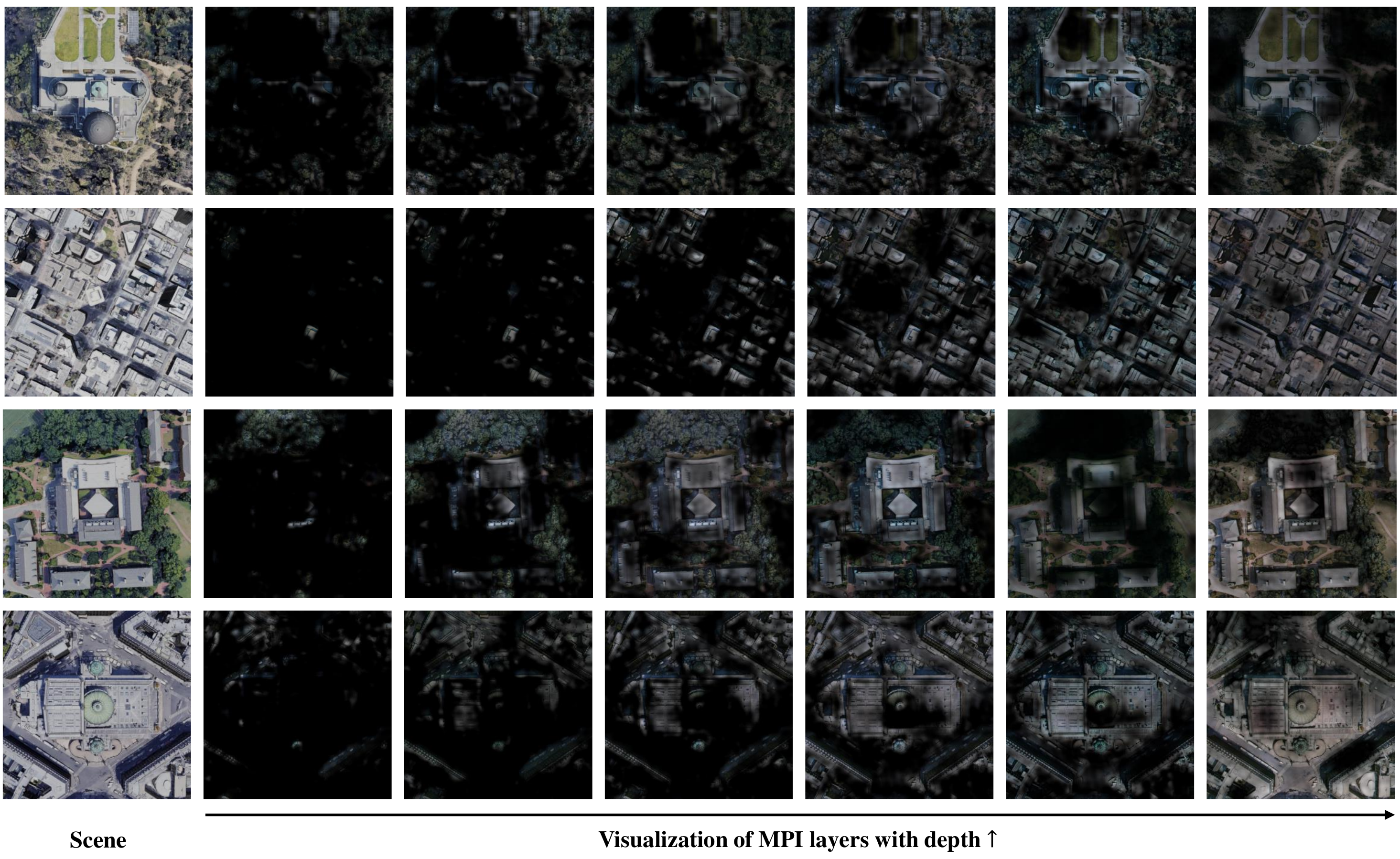}
    \caption{Visualization of MPI layers with different depths after the cross-scene initialization. For 6 of the 32 MPI layers per scene, the product of $C_{z_{i}}$ and $\sigma_{z_{i}}$ is visualized in this figure.}
    \label{fig:visualization_mpi}
\end{figure*}

\subsection{Controlled Experiment}
\label{ssec:controlled}

\textbf{Ablation study}. We perform ablation studies on 1) cross-scene initialization and 2) per-scene optimization to validate their effectiveness.

For cross-scene initialization, we remove this step from the pipeline and optimize the implicit MPI representation directly. For the input $F_{prior}$ of the MPI generator, we adopt two schemes: random initialization and setting as a learnable variable, corresponding to the second and third lines of Tab.~\ref{tab:necessity_analysis}. According to quantitative comparison results, the network needs more iterations to converge without the learning-based initialization step. Optimizing $F_{prior}$ and $\phi$ simultaneously brings better performance, but introduces more parameters at the same time. In comparison, ImMPI converges faster with learning-based initialization and reaches a higher accuracy. This indicates that the cross-scene initialization not only can speed up per-scene optimization, but regularize the optimization to avoid local minimum solutions.

As for the per-scene optimization, we directly render novel views from the ImMPI after the initialization. From the first line of Tab.~\ref{tab:necessity_analysis}, we can see all the three metrics decrease a lot when removing the per scene optimization step. Although there are no significant perspective changes in remote sensing scenes, a single image is still insufficient to support the novel view synthesis. Combined with multi-view information, per-scene optimization significantly improves the quality of rendered images.

\begin{table}
    \centering
    \caption{Ablation Study on Cross Scene Initialization (Init) and Per Scene Optimization (Optm)}
    \resizebox{0.5\textwidth}{!}{
    \begin{tabular}{cc|cc|ccc}
    \toprule
         Init & Optm & Iteration & Variable & PSNR$\uparrow$ & SSIM$\uparrow$ & LPIPS$\downarrow$ \\
    \midrule
        $\checkmark$ & $\times$ & - & - & 16.04 & 0.3990 & 0.4134 \\
        $\times$ & $\checkmark$ & 1000 & $\phi$ & 24.77 & 0.7568 & 0.2828 \\
        $\times$ & $\checkmark$ & 1000 & $\phi + F_{prior}$ & 24.81 & 0.7663 & 0.2698 \\
        $\checkmark$ & $\checkmark$ & 200 & $\phi$ & 24.89 & 0.7922 & 0.2082 \\
        $\checkmark$ & $\checkmark$ & 500 & $\phi$ & \textbf{25.95} & \textbf{0.8306} & \textbf{0.1734} \\
    \bottomrule
    \end{tabular}
    }
    \label{tab:necessity_analysis}
\end{table}

\textbf{Effect of depth hypothesis number}.

The accuracy of reconstruction result can be affected by the number of depth hypothesis $D$ in ImMPI. Here we quantitatively analyze the impact by experiment. Specifically, we construct explicit MPI representations with 8, 16, 32 and 48 layers produced by different depth hypothesis numbers in MPI generator. As $D$ increases, the parameters of ImMPI remain the same, but the memory and computation consumption during rendering increases significantly, reducing the speed of per-scene optimization. The results are shown in Tab.~\ref{tab:depth_ablation}: PSNR and SSIM increase when $D$ grows from 8 to 32, while the FLOPs almost quadruple increases. In the case of $D=48$, the quality of rendered image drops slightly. We attribute the decrease to overfitting to training views due to excessive depth sampling. The further increase of depth hypothesis number brings limited accuracy improvement, but the computational resource consumption cannot be ignored. Therefore, we finally set $D$ to 32 in our method.

\begin{table}
    \centering
    \caption{Comparison of memory usage, parameter amount, and computation consumption (FLOPs) when rendering an image with different depth hypothesis number $D$ in our method.}
    \resizebox{0.5\textwidth}{!}{
    \begin{tabular}{c|ccc|ccc}
    \toprule
         D & PSNR$\uparrow$ & SSIM$\uparrow$ & LPIPS$\downarrow$ & Memory$\downarrow$ & Params$\downarrow$ & FLOPs$\downarrow$ \\
    \midrule
        8 & 22.99 & 0.6419 & 0.3215 & 6981MB & 16.64M & 78.57G \\
        16 & 24.25 & 0.7419 & 0.2600 & 10774MB & 16.64M & 147.53G \\
        32 & \textbf{25.95} & \textbf{0.8306} & \textbf{0.1734} & 18446MB & 16.64M & 285.46G \\
        48 & 24.78 & 0.7869 & 0.2286 & 23872MB & 16.64M & 423.38G \\
    \bottomrule
    \end{tabular}
    }
    \label{tab:depth_ablation}
\end{table}

\textbf{Effect of training view number.} In this part we analyze the sensitivity of our method to the number of training views adopted during the per-scene optimization. For each scene in the LEVIR-NVS dataset, we experiment with 3, 5, 7, 9 and 11 training views for optimization. From Tab.~\ref{tab:trainview_ablation}, all three metrics in train-view and test-view increased with the increase of the number of training views. This shows that more perspectives prompt more accurate reconstruction. Note that ImMPI outputs satisfactory results even with only 3 training views, indicating our method can generalize well to very sparse views in remote sensing scenes.

\begin{table}
    \centering
    \caption{View synthesis accuracy with different number of training views. $N$ denotes the number of training views used during the per-scene optimization.}
    \resizebox{0.5\textwidth}{!}{
    \begin{tabular}{c|ccc|ccc}
    \toprule
        \multicolumn{1}{c|}{} & \multicolumn{3}{c|}{\textbf{Train-View}} & \multicolumn{3}{c}{\textbf{Test-View}} \\
        N & PSNR$\uparrow$ & SSIM$\uparrow$ & LPIPS$\downarrow$ & PSNR$\uparrow$ & SSIM$\uparrow$ & LPIPS$\downarrow$ \\
    \midrule
        3 & 23.32 & 0.7445 & 0.2294 & 23.16 & 0.7424 & 0.2304 \\
        5 & 24.66 & 0.8016 & 0.1973 & 24.70 & 0.8019 & 0.1971 \\
        7 & 25.46 & 0.8168 & 0.1850 & 25.53 & 0.8175 & 0.1843 \\
        9 & 25.95 & 0.8226 & 0.1794 & 25.56 & 0.8183 & 0.1812 \\
        11 & \textbf{26.34} & \textbf{0.8351} & \textbf{0.1717} & \textbf{25.95} & \textbf{0.8306} & \textbf{0.1734} \\
    \bottomrule
    \end{tabular}
    }
    \label{tab:trainview_ablation}
\end{table}

\subsection{MPI Visualization}
\label{ssec:MPI_visualization}

To explore the effect of the cross-scene initialization, we visualize the product of $C_{z_{i}}$ and $\sigma_{z_{i}}$ of several layers of initial MPI in Fig.~\ref{fig:visualization_mpi} for qualitative analysis. From the figure, it can be seen that ground objects such as buildings and trees of different heights in the same scene appear in different MPI layers. As the depth increases (away from the camera), the content in the scene from the roof to the ground gradually emerges. This indicates that our model can successfully learn depth information from a single image. Also, note that the Prior extractor is trained using the WHU MVS/Stereo dataset, while test image is from the LEVIR-NVS dataset. The visualization results show that our method can accommodate the domain gap between different datasets to some degree. Although the initial ImMPI has some errors in detail, it's sufficient as a good initialization to circumvent some suboptimal solutions. Meanwhile, the initialization plays as a regularization role in the optimization process, avoiding overfitting to training perspectives.

% This is similar to the height distribution, but not exactly the same. Cause the sigma of MPI referring to transparency does not corresponding to height/attitude physically. 
% , which means that the network has the ability to infer the scene geometry rather than just focusing on the specific content. 

\subsection{Depth Estimation}
\label{ssec:depth}

\begin{figure}[t]
    \centering
    \includegraphics[width=0.5\textwidth]{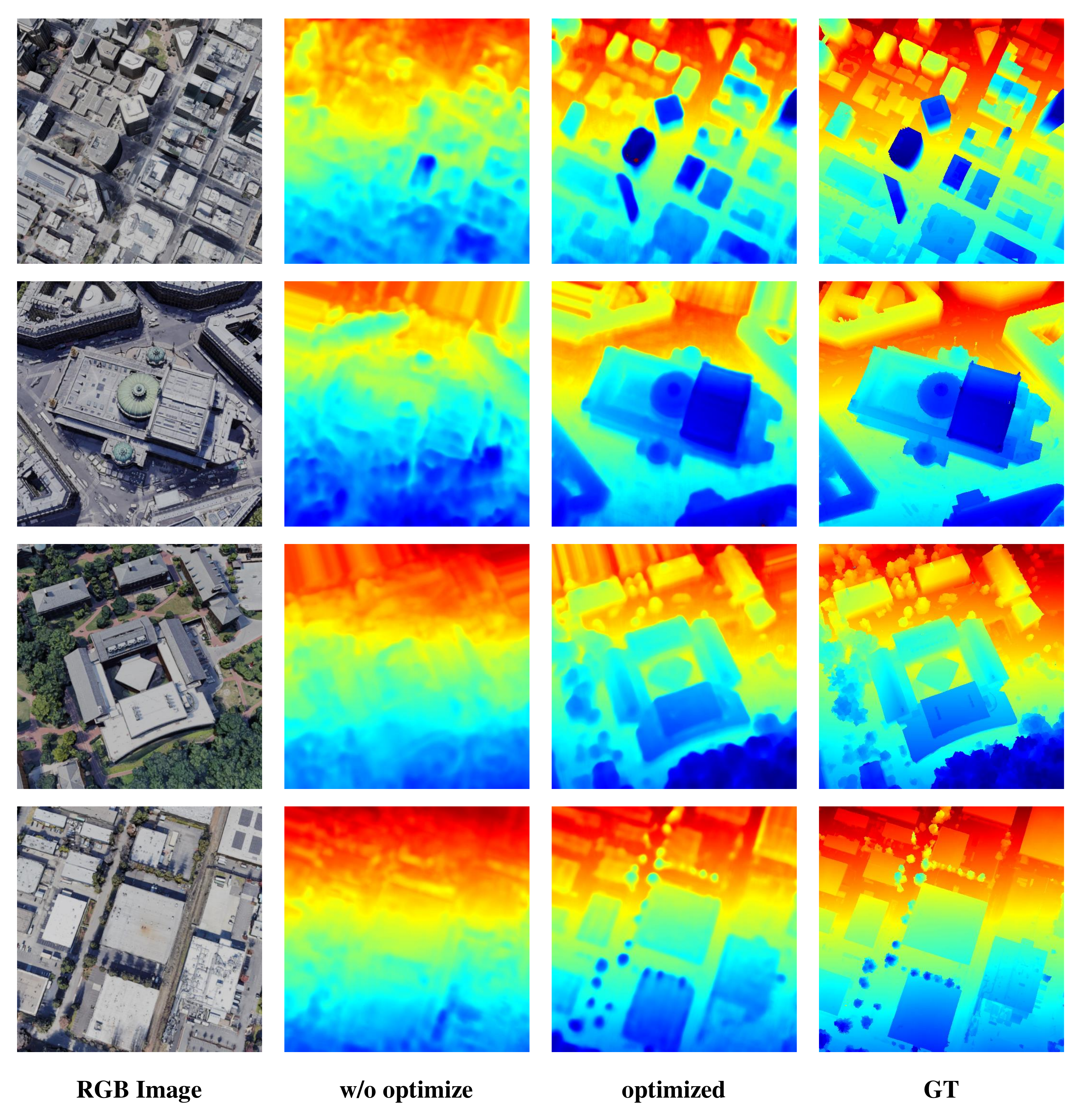}
    \caption{Visualization of the rendered depth map with our method. ``w/o optimize'' refers to the depth maps directly rendered from MPI with only initialization. ``optimized'' refers to the depth maps obtained after the per-scene optimization. Cool color means objects are closer to the camera while warm color means the opposite. }
    \label{fig:visualization_depth}
\end{figure}

In addition to novel view synthesis, our method can estimate the depth map of the corresponding view according to Equation~\ref{equ:depth}. Depth map qualitative results are shown in Fig.~\ref{fig:visualization_depth}. For each scene, we visualize one of the test view depth estimation results. As we can see from the figure, the MPI only with initialization can roughly predict the distance between the ground objects and the camera. Per-scene optimization introduces information from other perspectives, further improving depth estimation accuracy. Compared with the actual depth value, the depth map obtained from MPI still has errors at pixels where the depth value changes sharply. Nevertheless, since we mainly focus on the novel view synthesis task and the depth map is a by-product, our approach is still a feasible way for depth estimation.

%  ******************************Conclusion******************************
\section{Conclusion}
We propose a new method named ImMPI and a new dataset named LEVIR-NVS for remote sensing novel view synthesis. Given a set of images from a scene, novel view RGB images and corresponding depth maps can be rendered from optimized ImMPI by differentiable rendering. ImMPI combines the advantages of implicit neural network and explicit MPI representation and is naturally suitable for remote sensing images. The Implicit representation encodes the scene geometry to network weights with fewer parameters and the explicit MPI achieves faster rendering speed. We also propose a learning-based cross-scene initialization method to extract scene priors, which dramatically speeds up the per scene optimization and improves accuracy under sparse view inputs. Compared with NeRF-based methods, ImMPI shows significant improvement with 2 times optimization speed and 20 times rendering speed.

% Our empirical evidence indicates ImMPI is more efficient and effective than purely implicit neural representations. 

%  ******************************appendices******************************  
% \appendices

% % you can choose not to have a title for an appendix
% % if you want by leaving the argument blank
% \section{}

% %  ******************************Acknowledgment******************************   
% % use section* for Acknowledgment
% \section*{Acknowledgment}


\ifCLASSOPTIONcaptionsoff
  \newpage
\fi

%  ******************************references section******************************    
\bibliographystyle{IEEEtran}
\bibliography{references}

%  ******************************biography section ******************************    

\end{document}